\documentclass{bmvc2k}

\usepackage{amssymb}
\usepackage{times}
\usepackage{epsfig}
\usepackage{graphicx}
\usepackage{amsmath}
\usepackage{amssymb}
\usepackage{algorithmic}
\usepackage{algorithm}
\usepackage{memhfixc}
\usepackage{amstext}
\usepackage{stmaryrd}
\usepackage{fancyhdr}
\usepackage{booktabs}
\usepackage{url}
\usepackage{tabularx}
\usepackage[toc,page]{appendix}

\dblfloatsep\floatsep
\dbltextfloatsep\textfloatsep
\makeatletter
\renewcommand\normalsize{%
  \@setfontsize\normalsize\@xpt\@xiipt
  \abovedisplayskip 6\p@ \@plus2\p@ \@minus3\p@
  \abovedisplayshortskip \z@ \@plus3\p@
  \belowdisplayshortskip 4\p@ \@plus3\p@ \@minus3\p@
  \belowdisplayskip \abovedisplayskip
  \let\@listi\@listI}
\normalsize

\title{Learning a Robust Society of Tracking Parts}

\addauthor{Elena Burceanu}{eburceanu@bitdefender.com}{123}
\addauthor{Marius Leordeanu}{marius.leordeanu@imar.ro}{14}

\addinstitution{
 Institute of Mathematics of the\\
 Romanian Academy, Bucharest
}

\addinstitution{
 Bitdefender, Romania\\
}

\addinstitution{
 Department of Mathematics and Computer Science\\
 University of Bucharest, Romania\\
}

\addinstitution{
 Department of Automatic Control and Computer Science\\
 University Politehnica of Bucharest\\
}

\runninghead{E. Burceanu, M. Leordeanu}{Learning a Robust Society of Tracking Parts}

\begin{document}

\maketitle

\begin{abstract}
Object tracking is an essential task in computer vision that has been studied since the early days of the field. 
Being able to follow objects that undergo different transformations 
in the video sequence, including changes in scale, illumination, shape and occlusions, makes the problem extremely difficult.
One of the real challenges is to keep track of the changes in objects appearance and not drift towards the background clutter. Different from previous approaches, we obtain robustness against background with a tracker model that is composed of many different parts. They are classifiers that respond at different scales and locations. The tracker system functions as a society of parts, each having its own role and level of credibility. Reliable classifiers decide the tracker's next move, while newcomers are first monitored before gaining the necessary level of reliability to participate in the decision process. Some parts that loose their consistency are rejected, while others that show consistency for a sufficiently long time are promoted to permanent roles. The tracker system, as a whole, could also go through different phases, from the usual, normal functioning to states of weak agreement and even crisis. The tracker system has different governing rules in each state. What truly distinguishes our work from others is not necessarily the strength of individual tracking parts, but the way in which they work together and build a strong and robust organization. We also propose an efficient way to learn simultaneously many tracking parts, with a single closed-form formulation. We obtain a fast and robust tracker with state of the art performance on the challenging OTB50 dataset.
\end{abstract}

%-------------------------------------------------------------------------
\section{Introduction}
\label{sec:intro}

Object tracking is one of the first and most essential problems in computer vision. While it has attracted the interest of many researchers over several decades of computer vision, the task is far from being fully solved~\cite{VOT2014, MOT2014, OTB2015}. The problem is difficult for many reasons, including the severe changes in object appearance, presence of background clutter and occlusions that take place in 
the video sequence. Moreover, the only ground truth knowledge given to the tracker is the bounding box of the object in the first frame. Thus, without knowing in advance the properties of the object being tracked (neither the ones of a general object), the tracking algorithm must learn them on the fly. It must adapt correctly and make sure it does not drift toward other objects in the background.

Different from previous methods, our proposed tracking model is composed of a large group of different object part classifiers, which act together like a society. Each classifier takes care of a different part of the object, at a certain scale and location. It also has its own level of credibility.
The overall tracker is, on one hand, kept robust and stable through a group of reliable classifiers. At the same time, it can adapt to new conditions by considering new candidate part classifiers. Candidates are continuously being monitored and promoted or rejected based on their estimated reliability. The ability to learn a large group of classifiers efficiently, over the video sequence, is given by our proposed multi-class approach using regularized least squares that is based on a novel theoretical insight, presented in detail in Section \ref{sec:math}. \\

\noindent \textbf{Relation to previous work:}
There are many tracking methods which differ mainly in terms of
target region, appearance model, mathematical formulation and optimization. Objects can be represented by boxes, ellipses~\cite{Kuo2011HowDP}, superpixels~\cite{Wang2011SuperpixelT} or blobs~\cite{Godec2011HoughbasedTO}. The appearance model can be described
as one feature set over the region or an array of features, one for each part of the target~\cite{Forsyth2010ObjectDW,Shu2012PartbasedMT,KwonL09}.
Part models are more resistant to occlusions and non-rigid appearance changes. 

Features used by tracking models could be either simple raw pixel information or more specialized ones that describe regions and keypoints, which could better handle changes in viewpoint, scale, illumination and deformations. Such features include Gabor \cite{Nguyen2006RobustTU}, HOG \cite{Forsyth2010ObjectDW}, SIFT, SURF \cite{Chu2010ColorIS}, Haar \cite{MIL2011}, or combinations of low and high level features \cite{Ma2015HierarchicalCF} from widely used, pre-trained CNNs. Some recent algorithms start applying powerful features and classifiers pre-learned with deep convolutional networks (CNNs)~\cite{nam2016learning,danelljan2016beyond,chen2016convolutional,ma2015hierarchical,qi2016hedged}.The disadvantage of current methods using CNNs is the need to learn in advance the features used on large, human labeled, datasets. They are valid only on objects with the same particularities like the ones in the dataset. 

Both the classifier and the features are known to be very important for classification performance.
This can also be observed in the recent method proposed in~\cite{DBLP:journals/pami/HareGSVCHT16}, where only the hard negatives and positives are kept for the tracker model based on SVM. 
When using more complex features, the performance of their tracker increases by up to 10\%.
Some methods augment the features with information from optical flow \cite{Brendel2011MultiobjectTA, Kalal2012Pami}, segmentation \cite{Godec2011HoughbasedTO} or superpixels \cite{Wang2011SuperpixelT}. More complex methods use active appearance models \cite{Hu2013ActiveCV}. 
 
Besides accuracy, speed is also very important in tracking. Ideally, the tracker should operate in real-time. One of the fastest methods for tracking uses correlation filters~\cite{DBLP:conf/cvpr/BolmeBDL10}. More recent work in this direction formulates the problem using circular matrices, easy to decompose in the FFT domain and split it in independent equations, with closed-form solutions\cite{HenriquesCMB14}. The immediate advantage is speed, but the resulting, elegant tracker is also very competitive, while being an order of magnitude faster than its competitors~\cite{OTB2015}. Again, the best performance is obtained when more complex features are used (HOG) than simple raw pixel values.

In relation to previous work, our model uses many tracking parts (about $100-600$), which are learned fast and simultaneously, in a given frame. We propose a novel formulation in the context of tracking based on regularized least squares. We use only very simple features, based on raw pixel values, our strength being based entirely on the model that functions as a robust society of many parts. Note that our model is general and can accommodate the use of any combination of features and classifiers for the separate object parts. In brief, our main contributions are: \\

\noindent \textbf{Main contributions:} 1) Our first contribution, discussed in Sections \ref{sec:society} and \ref{sec:algo} is the
concept and design of the overall tracker that functions as a robust society of many different classifier parts, at different locations and scales and with different weights and reliability levels. Thus, our system is able to keep its stability over time, while also adapting to the current changes of the object in the video. On the difficult OTB50~\cite{WuLimYang13} dataset it outperforms by a significant margin current state-of-the-art methods that do not use CNN features pre-trained on large human labeled datasets.
2) Our second contribution enables the efficient implementation of the tracker. We are able to learn simultaneously many part classifiers using a novel weighted one-vs-all regularized least squares formulation, with closed-form solution and important theoretical properties, as discussed in Section \ref{sec:math}. 

\section{Intuition and motivation}
\label{sec:society}

Visual tracking is about being able to adapt the current knowledge about the object model
to changes that take place continuously in the stream of video. How could the tracker learn 
novel aspects of the object of interest and, at the same time, not forget valuable older information?
Most current learning methods that continuously adapt to new information could slowly forget the initial models they started from - and those initial models could still be valid and useful for future use.

We argue that a tracking model composed of many parts, each with its own degree of reliability (or trust), which function together according to certain rules that consider their different roles and specific trust levels, could have the two highly desirable properties. The tracker, functioning like a society of tracking parts, could be both stable in the face of rapid and noisy variations in the environment and could also adapt and learn when meaningful changes take place. 

We draw an analogy between the model we propose and a simplified form of a human community (or organization), in which
people have different roles and degree of importance. Certain people, very few, are the founders of that community. They are very often considered reliable, from the start. 
Over the longer term, the community is ruled by the shared responsibility of a group of reliable members, who include
some of the initial founders and those who have proved their credibility of the time. 
In our case, these members would be responsible for deciding the next tracker move. At the lower level, the organization is continuously refreshed with newcomers, young members who want to become part of the core group of leaders, but are not yet ready to rule.  While they provide a constant source of new and potentially beneficial information that could be better suited to current changes in the "world", their
consistency is not yet proven. New members are first being monitored, without being allowed to make decisions that could affect the behaviour of the whole community. Once they prove their value they are moved to the core of reliable members with decision power. At the same time, current members could loose credibility if they stop showing consistency. Those will eventually be rejected. Others, who have proven reliability for long enough, are promoted to a special permanent member status.

In Section \ref{sec:algo} we explain in detail how one could measure reliability for tracking parts. In brief, we consider a part to be reliable if it has showed independently and frequently enough agreement in voting with the majority of the other parts. Since the majority is statistically robust, the estimation of reliability in this way is also robust. 

By considering members with different capabilities and roles, in many ways similar to a human organization, the tracker becomes a system that could display the following important properties: \\

\noindent \textbf{1) Stability:} the core members sustain constant reliable functioning. They act independently and decide by majority, providing robustness against noisy variations.
Only the simple and the permanent (gold) reliable members can influence the 
majority vote for the next tracker move.\\

\noindent \textbf{2) Adaptation:} the tracker is able to continuously adapt by adding new trackers and removing old ones, as time passes. It promotes the new reliable members and eliminates the ones that lost reliability (excepting the permanent members). Note that gaining and loosing reliability can happen only over time. It is the temporal buffer, when tracking parts are monitored, which ensures both stability and the capacity to adapt to new conditions.\\

\noindent \textbf{3)Ability to never forget:} Tracking parts that display consistent reliable behaviour over longer periods of time are promoted to the status of permanent or gold members. Thus we ensure that the model does not forget information that has been proven consistent and could be of vital importance in the future. \\

\section{Algorithm}
\label{sec:algo}

\begin{figure}[t]
\begin{center}
   \includegraphics[width=1\linewidth]{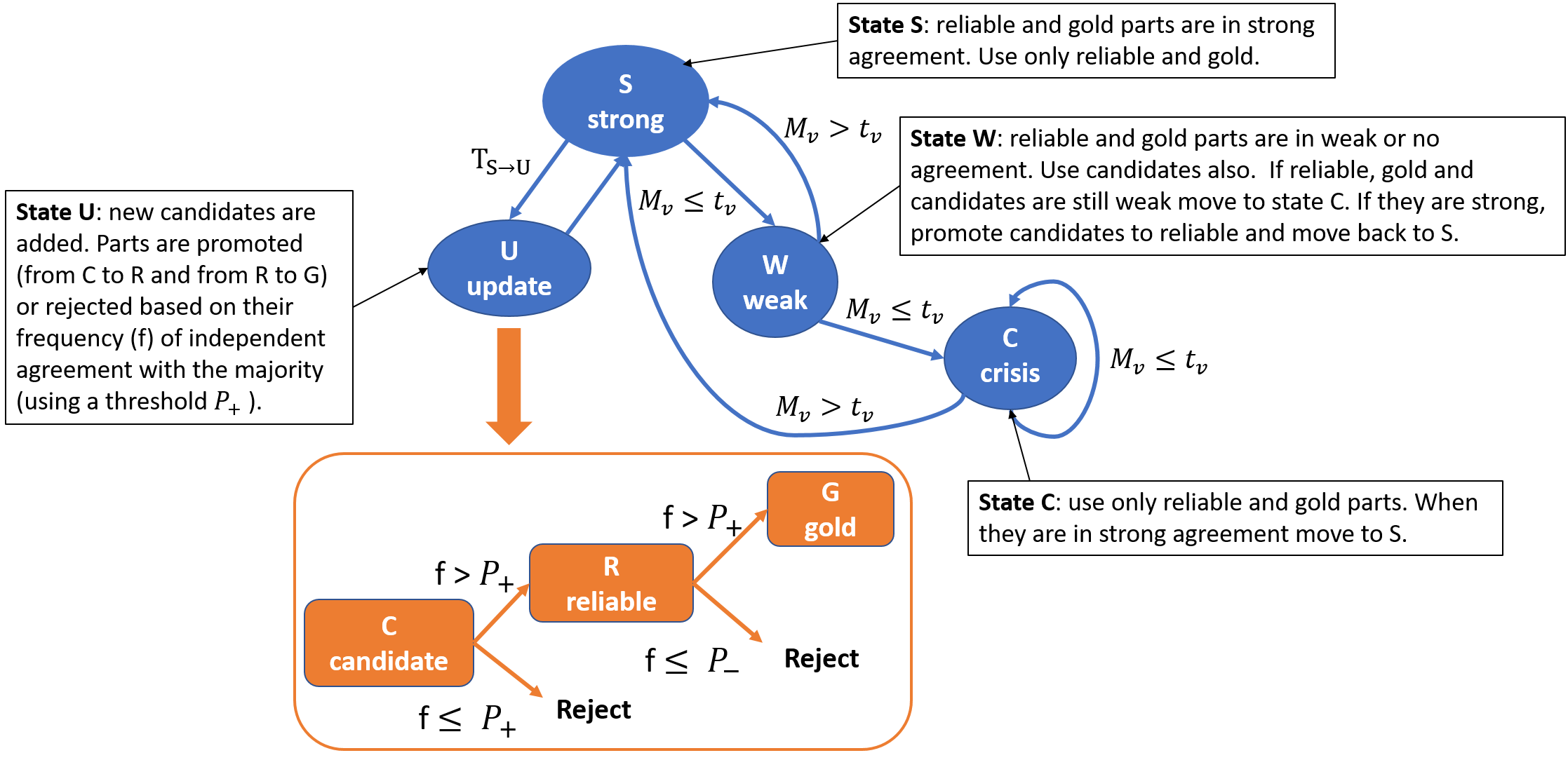}
\end{center}
   \caption{Overview of our approach: the tracker, as a society of parts
   can transition between different states, depending on the difficulty of the situation. In each state, parts contribute in different way to deciding the next move. The tracking parts themselves, also have different states, depending on their reliability, which is learned over time. Thus, the tracker functions as a robust system.}
\label{fig:STP}
\end{figure}

The proposed tracker algorithm, 
which we term Society of Tracking Parts (STP), is based on 
a system of part classifiers (Figure \ref{fig:STP}). The tracker is learned and formed online, during tracking, from scratch, starting from the first, ground truth bounding box. \\

\noindent \textbf{Tracking by voting:}
The tracker always chooses as its next move at time $t$, 
the place (the center of the bounding box) $l_{t+1}$
where there is the largest accumulation (of value $M_v$) of parts votes
, within a certain region $R_t$. 
This searching zone for the target is restricted around the previous bounding box, over a region defined by a given parameter $\delta$. For each part $i$ there is an activation map $A_{ti}$, computed as the response of the classifier $c_i$ corresponding to that part over the search region $R_t$. 
The activation maps of considered parts are each shifted with the part displacement from object center and added together to form the overall $A_t$. $A_t$ is the voting map for the center and when parts are in strong agreement, all votes focus around a point (the next predicted bounding box center).
After smoothing $A_t$ with a small Gaussian filter, the maximum is chosen as the next center location $l_{t+1}$. 
Note that different parts are allowed to contribute with their activation maps, depending on their reliability and tracker state (see also Figure \ref{fig:STP}), as described next. \\

\noindent \textbf{Part reliability states:}
Reliability of a part $i$ is estimated as the frequency $f_i$ at which the maximum activation of a given part is in the neighborhood (within 5 pixels in our implementation) of the maximum sum activation where the next tracker center $l_{t+t}$ is chosen.
If a part is selected for the first time, it is considered a candidate part. Every $T_{S->U}$ frames, the tracker measures the reliability of a given part, and promotes parts with a reliability
larger than a threshold $f_i > p_+$, from candidate state (C) to reliable state (R) and from reliable (R) to gold (G) (Figure \ref{fig:STP}). Parts that do not pass the test $f_i \leq p_-$
are removed, except for gold ones which are permanent. Tracker states are: \\

\noindent \textbf{Tracker states.}
Strong (S) - in the "strong" (S) state the tracker is ruled by the voting of the reliable and gold parts. When the maximum over the sum of their activation maps is over a threshold ($M_v > t_v$), 
tracking is considered strong.
Every $T_{S->U}$ frames the tracker enters the "update" (U) state  from the S state. \\
Update (U) - in the U state, the tracker considers new classifiers from the current frame as candidates, learned from patches that cover areas of the bounding box where current reliable and gold members have weak responses. The new candidates will be monitored from then on and their reliability will be estimated, based on their consensus frequency with the weighted majority, as discussed previously. Candidate votes are not taken in consideration until they become "reliable" parts. In this state, existent parts (candidates and 
reliable) are promoted or rejected, based on their reliability $f_i$ as also discussed previously. \\
Weak (W) - when the maximum accumulating vote ($M_v$) in the S state is weak ($M_v \leq t_v$) the tracker enters the W state. In this state, candidates from previous (strong) states are allowed to vote together with the reliable and gold members. If the total accumulation $M_v$ is still weak, the tracker enters the state of "crisis" (C). Otherwise it promotes to "reliable" the candidates that agreed with the majority vote, then goes back to S state. \\
Crisis(C) - state C is enetered from W, when the votes in W are weak. In C,
the tracker starts searching the entire image (basically, region $R_t$ becomes the entire image). Then it moves to
to the maximum accumulation of the reliable and gold members. When $M_v > t_v$ it goes back to the strong state S. Until then it stays in C, with no member update allowed.

\begin{figure}[t]
\begin{center}
   \includegraphics[width=1\linewidth]{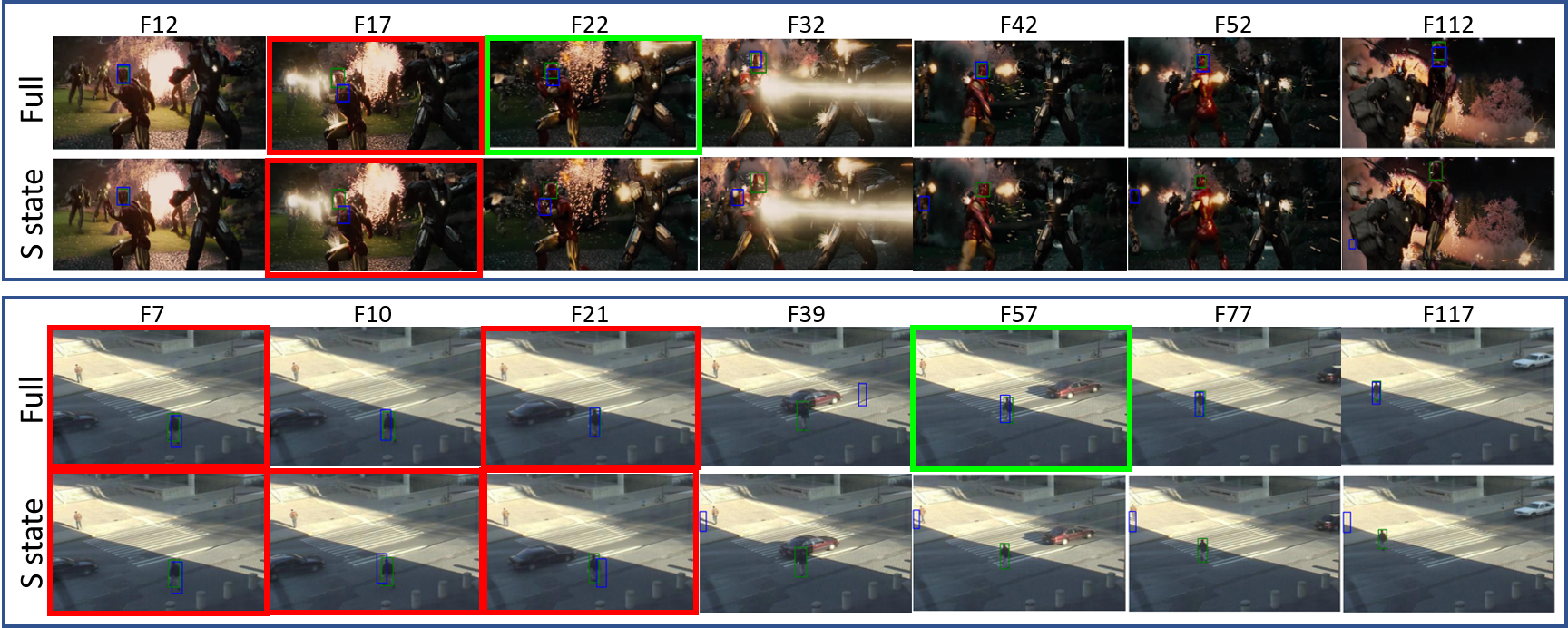}
\end{center}
   \caption{Qualitative differences in tracker performance, when different tracker states are allowed. "full" corresponds to the tracker with all states allowed, while "S state" has only S allowed. Red frames correspond to frames with weak votes. Green frames are frames where the tracker recovered after being lost. Note the better performance of the full tracker.
   }
\label{fig:states}
\end{figure}

\textbf{Example:} In Figure \ref{fig:states} we show 
qualitative results to demonstrate the importance of the different tracker states. In the Iron Man sequence (top), the "full" tracker (with all states activated) stays with the main object until the end of the sequence and recovers from moments when it is lost. The "S-only" tracker (with states W and C deactivated) once lost, remains lost. In the second Crossing example, both versions of the tracker have weak votes at frame 7. The full tracker enters the weak state and recovers in a better position, while having promoted new candidates to reliable in the W state. Around frame F21 both trackers are lost, but during Crisis, the full version recovers in frame F57 and stays with the person crossing until the end, unlike the S-only tracker that is lost from F21. Note that in our experiments, we show in Table \ref{tab:states} quantitative differences between versions of the tracker with different state subsets allowed (S, SW and all SWC), which fully justify the use of all three states. \\

\noindent \textbf{Learning the tracker:}
The mathematical details related to training the individual classifiers 
are discussed in Section \ref{sec:math}. 
In order to keep the appearance model up to date, in the update phase STP chooses new patches to add as positive parts. Only patch classifiers that are highly discriminative from the rest are selected. A patch classifier is considered discriminative if the ratio between the response on the positive patch (its own corresponding patch)
and the maximum response over negatives is larger than a threshold $t_d$. Positives are selected from the inside of the bounding box, while (hard) negatives are selected as patches from outside 
regions with high density of edges.
We sample patches from a dense grid (2 pixels stride) of small (17x17), medium (27x27) and full bounding box sizes. The small ones will see local appearance, and the larger ones will contain some context. A point in grid is covered only by one selected discriminative patch, at one size. 
The smaller ones have priority and we search the next size for the patch centered in the grid point only if the smaller patch is not discriminative enough. The object box is covered when each pixel is covered by any given patch.
A simple budgeting mechanism is added, in order to limit the speed impact. When too many parts of a certain patch size become reliable $>N_{max}$, we remove the new reliable ones which are most similar to older parts, based on simple dot product similarity for the classifiers.

\textbf{Parameters:} we use the following parameters values in our experiments from Section \ref{sec:experiments}: $\delta = 25px$, $T_{S->U}=10$ frames, $t_d=1.4$, $p_+ = 0.2$, $p_- = 0.1$ and $N_{max} = 200$ parts.

\section{Mathematical aspects for learning the parts}
\label{sec:math}

We introduce the mathematical formulation for learning the classifiers for the tracking parts. For a given feature type let $\mathbf{d}_i \in \mathbb{R}^{1 \times k}$ be the $i$-th descriptor, with $k$ real elements, corresponding to an image patch window at a certain scale and location relative to the object bounding box. Note that in our experiments, the features we use are simple pixel values from seven image channels, the three color channels, plus four more channels representing the gradient magnitudes over four orientations $(0, \pi/4, \pi/2, 3\pi/4)$ 
The descriptor $\mathbf{d}_i$ is
a vectorised version of the specific patch concatenated over all image channels.
Let $\mathbf{D}$ be the data matrix, formed by putting all descriptors in the image one row below the other. 

We learn the optimal linear classifier $\mathbf{c}_i$ that separates $\mathbf{d}_i$ from the rest of the patches, according to a regularized linear least squares cost -  which is both fast and accurate. Classifier $\mathbf{c}_i$ minimizes the following cost (\cite{DBLP:books/lib/Murphy12} Ch. 7.5):

\begin{equation}
\label{eq:learning_cost}
    \min \frac{1}{n}\|\mathbf{Dc}_i-\mathbf{y}_i\|^2 + \lambda \mathbf{c}_i^\top \mathbf{c}_i.
\end{equation}

In classification tasks the number of positives and negatives should be properly balanced, according to their prior distributions and the specific classifier used.
Different proportions usually lead to different classifiers. In linear least squares formulations 
weighting differently the data samples could balance learning. 

\paragraph{One sample versus all}

The idea of training one classifier for a single positively labeled data sample has been successfully used before, for example, in the context of training SVMs~\cite{malisiewicz2011ensemble}.
Normally, when using very few positive samples for training a ridge regression classifier, weighting is applied to balance the data. Otherwise the classifier response on the positive samples is too low. 
Here we show that when a single positive sample is used, weighting does not change the direction of the resulting classifier, even though it changes its magnitude. This makes it possible to easily normalize classifiers trained with different positive to negative ratios.

\noindent \textbf{Property 1:} for any positive weight $w_i$ given to the positive $i$-th sample, when the negative labels considered are $0$ and the positive label is $1$ and all negatives have the same weight $1$, the solution vector to the weighted least squares version of
Eq. \ref{eq:learning_cost}
will have the same direction (it might differ only in magnitude). In other words, it is invariant under L2 normalization. 

\noindent \textbf{Proof:}
Let $\mathbf{c}_i$ be the solution to Eq. \ref{eq:learning_cost}. 
At the optimum the gradient vanishes, thus the solution respects the following equality $(\mathbf{D^\top D} + \lambda \mathbf{I}_k)\mathbf{c}_i = \mathbf{D^\top}\mathbf{y}_i$. Since $y_i(i)=1$ and $y_i(j)=0$ for $j \neq i$, it follows that $(\mathbf{D^\top D} + \lambda \mathbf{I}_k)\mathbf{c}_i = \mathbf{d}_i$. Since the problem is convex, with a unique optimum,
a point that obeys such an equality must be the solution. 
In the weighted case, a diagonal weight $n \times n$ matrix $\mathbf{W}$ is defined, with different weights on the diagonal $w_{j}=\mathbf{W}(j,j)$, one for each data sample. In that case, the objective cost optimization in Eq. \ref{eq:learning_cost} becomes: 

\begin{equation}
\label{eq:weighted_learning_cost}
\min \frac{1}{n}\|\mathbf{W}^{\frac{1}{2}}(\mathbf{Dc}_i-\mathbf{y}_i)\|^2 + \lambda \mathbf{c}_i^\top \mathbf{c}_i.
\end{equation}

We consider when all negative samples have weight $1$ and the positive one is given $w_i$. 
Now we show that for any $w_i$, if $\mathbf{c}_i$ is an optimum of Eq. \ref{eq:learning_cost} then there is a real number $q$ such that $q\mathbf{c}_i$ is the solution of the weighted case. The scalar $q$ exists if it satisfies $(\mathbf{D^\top D} + \mathbf{d}_i \mathbf{d}_i^\top (w_i-1)+\lambda \mathbf{I}_k)q\mathbf{c}_i=w_i\mathbf{d}_i$. And, indeed, it can be verified that $q = \frac{w_i}{1+(w_i-1)(\mathbf{d}_i^\top \mathbf{c}_i)}$ satisfies the required equality. See Appendix \ref{subsec:lin_ridge_regress} for a detailed proof.

\paragraph{Efficient multi-class ridge regression}

The fact that the classifier vector direction is invariant under different weighting of the positive sample suggests that training with a single positive sample will provide a robust and stable separator. The classifier can be re-scaled to obtain values close to $1$ for the positive samples. 

Property 1 also 
indicates that we could compute classifiers for all positive patches in the bounding box at once, by using a single data matrix $\mathbf{D}$. We form the 
target output matrix $\mathbf{Y}$, with one target labels column $\mathbf{y}_i$ for each corresponding sample $\mathbf{d}_i$. Note that $\mathbf{Y}$ is, in fact, the $n \times n$ $\mathbf{I}_n$ identity matrix. We now write the multi-class case of the ridge regression model and finally obtain the matrix of one versus all classifiers, with one column classifier for each tracking part: $\mathbf{C} = \mathbf(D^\top D + \lambda \mathbf{I}_k)^{-1} \mathbf{D}^\top$.
Note that $\mathbf{C}$ is a regularized pseudo-inverse of $\mathbf{D}$. $\mathbf{D}$ contains one patch descriptor per line. In our case, the descriptor length is larger than the number of positive and negative samples, so we use the Matrix Inversion Lemma~\cite{DBLP:books/lib/Murphy12}(Ch. 14.4.3.2) and compute $\mathbf{C}$ in an equivalent form (see more in Appendix \ref{subsec:matrix_inversion_lemma}): 

\begin{equation}
   \mathbf{C} = \mathbf{D}^\top (\mathbf{D D^\top} + \lambda \mathbf{I}_n)^{-1} . 
\end{equation}

Now the matrix to be inverted is significantly smaller ($n \times n$ instead of $k \times k$).

\section{Experimental Analysis}
\label{sec:experiments}

We have evaluated our tracker on the challenging OTB50 dataset. It contains 50 difficult video sequences, combining a variety of videos with complex scenarios, grouped into different categories of difficulty, such as: illumination variation (IV), scale variation (SV), occlusion (OCC), deformation (DEF), motion blur (MB), fast motion (FM), in-plane rotation (IPR), out-of-plane rotation/in-plane rotation (OPR/IPR), out-of-view (OV), background clutter (BC), low resolution (LR). We compared our method against top tracking methods: KCF~\cite{DBLP:journals/pami/HenriquesC0B15},
STRUCK~\cite{DBLP:journals/pami/HareGSVCHT16},
TLD~\cite{DBLP:journals/pami/KalalMM12},ORIA~\cite{DBLP:conf/cvpr/WuSL12},
MIL~\cite{DBLP:journals/pami/BabenkoYB11},
MOSSE~\cite{DBLP:conf/cvpr/BolmeBDL10},
CT~\cite{DBLP:conf/eccv/Zhang0Y12}.
They all use, as ground truth information 
only the initial bounding box provided in the first frame and do not employ any pre-trained CNN features or object detectors.

We followed the same evaluation protocol as in~\cite{DBLP:journals/pami/HenriquesC0B15} \cite{DBLP:journals/pami/HareGSVCHT16} \cite{HenriquesCMB14} \cite{MIL2011} \cite{OTB2015} by considering 
the predicted target correct if its center is within a threshold distance from the ground truth and compute the average precision (per category and for the whole dataset). We choose the same threshold (20px) as ~\cite{DBLP:journals/pami/HenriquesC0B15}. At this threshold
the relative order between the compared trackers stabilizes. In Table \ref{tab:categories} we present the detailed results for each category. Our algorithm outperforms the current state of the art methods by
by a large margin, while using only very simple, pixel level features. Our closest competition uses both strong features and stronger models: Struck~\cite{DBLP:journals/pami/HareGSVCHT16} uses Haar and histogram features combined with various kernels and KCF~\cite{DBLP:journals/pami/HenriquesC0B15} uses HOG descriptors, also combined with a non-linear kernel). This concludes that the power of the method is into our algorithm, STP, that uses many weak classifiers acting together. 
The majority turns out to be superior to a well selected elite (fewer, but smarter classifiers).

\begin{table}
	    \setlength\tabcolsep{2pt} % default value: 6pt
		\begin{tabular}{|c|c|c|c|c |c|c|c|c|c |c|c|c|}
			\hline
			Algorithm &  OPR  & MB & BC & OCC  & SV & IV & LR & OV& FM&    DEF& All& FPS\\
		
			\hline
			\textbf{OURS (STP)} & \color{red}75.9 & \color{red}72.5 & \color{blue}68.4 & \color{red}78.5 & \color{red}76.9 & \color{red}73.2& \color{red}63.7& \color{red}73.4& \color{red}69.8 & \color{red}80.1 & \color{red}78.7 & 30 \\
			\hline
            KCF on HOG \cite{DBLP:journals/pami/HenriquesC0B15} & \color{blue}72.9 &\color{blue}65 & \color{red}73.3& \color{blue}74.9&\color{blue} 67.9& \color{blue}71.1& 38.1 & \color{blue} 65 & \color{green}60.2 &\color{blue} 74 & \color{blue}73.2 & \color{blue}172\\
			\hline
			
			Struck \cite{DBLP:journals/pami/HareGSVCHT16}& \color{green}59.7 & \color{green}55.1 & \color{green}58.5 & \color{green}56.4 & \color{green}63.9 & \color{green}55.8 & \color{blue} 54.5 & 53.9 & \color{blue}60.4 & \color{green}52.1 & \color{green}65.6& 20 \\
			
			\hline
			\textbf{KCF on pixels} \cite{DBLP:journals/pami/HenriquesC0B15}  & 54.1 & 39.4  & 50.3 & 50.5 & 49.2 & 44.8 & \color{green}39.6 & 35.8 & 44.1 & 48 & 56 & \color{green}154 \\
            \hline
			TLD \cite{DBLP:journals/pami/KalalMM12} & 59.6 &51.8 &42.8 & 56.3 & 60.6& 53.7& 34.9& \color{green}57.6 &55.1  & 51.2 & 60.8 & 28\\
            
            \hline
            ORIA \cite{DBLP:conf/cvpr/WuSL12}  &49.3 &23.4 & 38.9& 43.5& 44.5& 42.1& 19.5 & 31.5 & 27.4 & 35.5 & 45.7 & 9\\
			\hline
			MIL \cite{DBLP:journals/pami/BabenkoYB11}   & 46.6 & 35.7 & 45.6 & 42.7& 47.1 & 34.9 & 17.1 & 39.3 & 39.6 & 45.5 & 47.5 & 38\\
			\hline            
			\textbf{MOSSE} \cite{DBLP:conf/cvpr/BolmeBDL10} & 39 & 24.4 & 33.9 & 39.7 & 38.7 & 37.5 & 23.9 & 22.6 & 21.3 & 36.7 & 43.1 & \color{red}615 \\
			\hline            
			CT \cite{DBLP:conf/eccv/Zhang0Y12} &39.4& 30.6 & 33.9 &41.2 & 44.8 &35.9 & 15.2& 33.6& 32.3 & 43.5 & 40.6 & 64\\
            \hline            
		\end{tabular}
	
	\caption{Mean Precision percentage (at 20px threshold) for video categories in OTB50. Bold entries are using only pixel level features. The first place, in each category, is shown in red, the second in blue and the third in green. We are in first place in 9 out of 10 categories. On average (per video over dataset) we outperform all other methods by at least $5.5\%$.} 
	\label{tab:categories}
\end{table}

\begin{table}
	\begin{center}
		\begin{tabular}{|c|c|c|c|}
			\hline
			 Algorithm & STP-S & STP-SW & STP-full \\
			
 			\hline
			Precision (20px) & 63.97 & 70.48 & \textbf{78.7} \\
			\hline
		\end{tabular}
	\end{center}
	\caption{Gain in mean precision (in $\%$) on OTB50 from adding tracker states to STP that take into account the problem of weak vote accumulation. STP-S is STP functioning only with the S state, while STP-SW has both S and W. STP-full has all three S, W and C states.}

	\label{tab:states}
\end{table}

\noindent \textbf{Relative importance of different tracker states:}
We tested the performance of our tracker when not all states are allowed, in order to better understand the importance of handling differently difficult scenarios when the accumulation of votes is weak. They usually correspond to cases when the tracker needs to use its candidates (W state, when reliable parts could have become obsolete) or when it undergoes occlusions or severe appearance changes (C state). In Table \ref{tab:states} and Figure \ref{fig:states} (also discussed in Section \ref{sec:algo})
we present quantitative and qualitative results of these tests. The results clearly show that using all three states (S,W,C) is superior to the other versions.

\begin{figure}[t]
\begin{center}
   \includegraphics[width=1\linewidth]{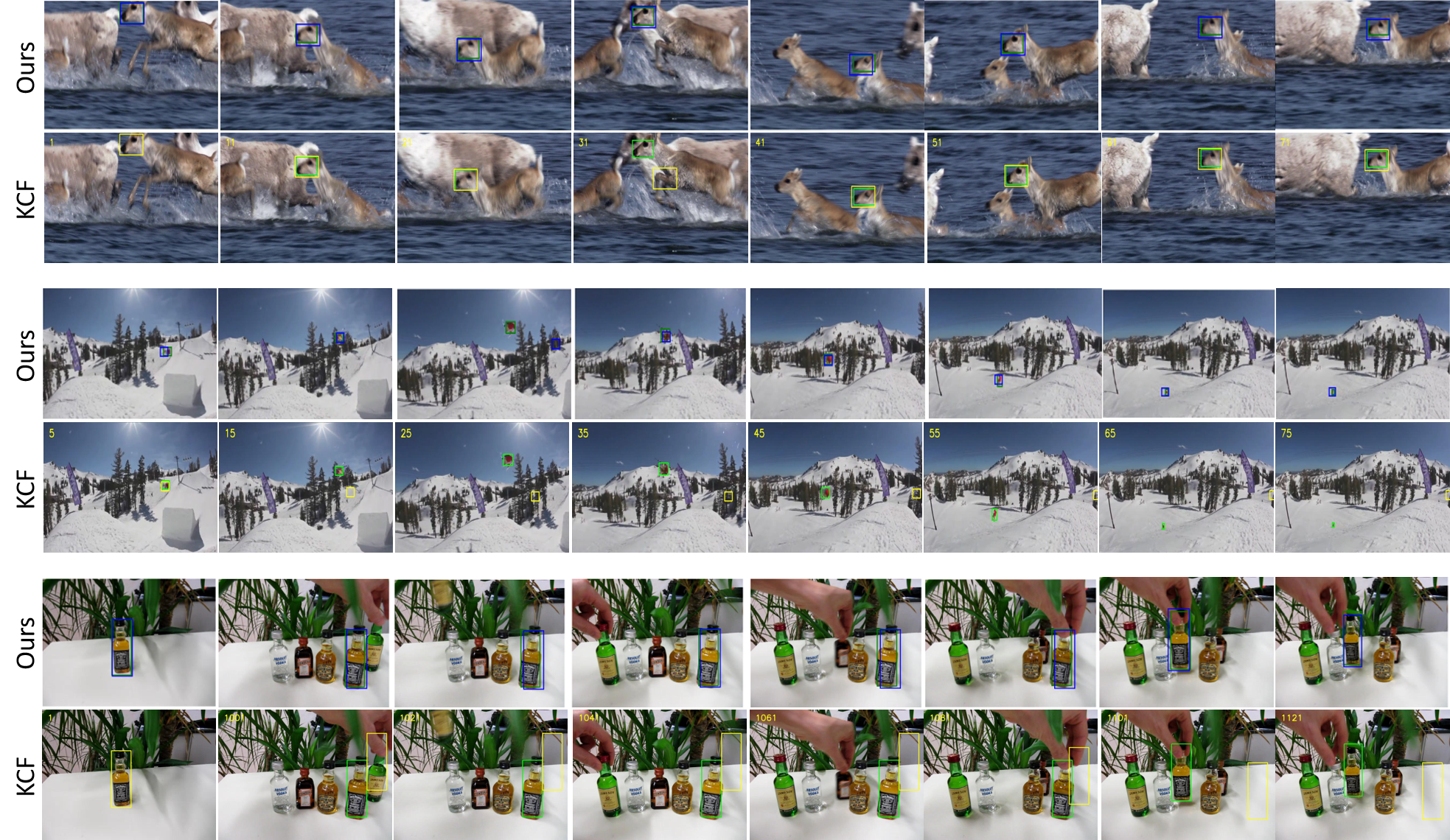}
\end{center}
   \caption{Qualitative performance of our method vs. the state of the art method KCF on some difficult cases. Our method handles very well many challenging scenarios.
   }
\label{fig:ours_vs_KCF}
\end{figure}

\section{Conclusions}
\label{sec:concl}

We have presented a novel model for object tracking based on a society of tracking part classifiers that is robust to different challenges and achieves top results on a very difficult recent benchmark. The strongest advantage of our approach is its ability to learn and adapt its model online while also keeping it robust and stable. The reason this is possible is due to the many different parts learned, at different scales and locations with respect to the object and having different roles according to their levels of credibility. The tracker itself also has different states of functioning. During normal functioning, updates are done at regular time  intervals. At moments when agreement between established trackers is week, the candidates are given the chance to jump in. If that does not work either, the classifier enters a state of crisis with no more updates being allowed, until consensus is found again. The tracker functions as an organization, with many different parts that may be weak by themselves, but they are strong when acting in combination. We have proved the power of this idea on a challenging dataset, demonstrating state of the art performance against top methods in the field.

\paragraph{Acknowledgements:} This work was supported in part by UEFISCDI,
under project PN-III-P4-ID-ERC-2016-0007.

\bibliography{egbib}

\begin{appendices}

\section{Fast One-sample vs. All Ridge Regression}
\label{subsec:lin_ridge_regress}

It is easy to obtain the closed form solution for $c_i$, from linear ridge regression formulation (\cite{DBLP:books/lib/Murphy12} Ch. 7.5), by minimizing the convex cost $\frac{1}{n}\|\mathbf{Dc}_i-\mathbf{y}_i\| + \lambda \mathbf{c}_i^\top \mathbf{c}_i$, we get Eq. \ref{eq:lin_ridge_reg}. It results the well known solution by inverting the positive definite matrix $\mathbf{D^\top D} +\lambda \mathbf{I}_k$.

\begin{equation}
    \label{eq:lin_ridge_reg}
   (\mathbf{D^\top D} +\lambda \mathbf{I}_k) \mathbf{c}_i = \mathbf{D^\top} \mathbf{y}_i
\end{equation}

In "one vs all" context, we choose 
$y_i^\top$ =
$
\begin{matrix}
        0 & 0 & ... & 1 & ... & 0 & 0\\
\end{matrix}$, with 1 only on the $i^{th}$ position. So, the multiplication with $y_i$ selects a column form $\mathbf{D}$: $\mathbf{D^\top} \mathbf{y}_i = \mathbf{d}_i$. Eq. \ref{eq:lin_ridge_reg} becomes:

\begin{equation}
    \label{eq:simplified_lin_ridge_reg}
    (\mathbf{D^\top D} +\lambda \mathbf{I}_k) \mathbf{c}_i  = \mathbf{d}_i
\end{equation}

When building classifiers, the classes should be balanced as numbers of entries. This ensures that the comparison between the activation scores for two different classifiers is valid. In "one vs all", usually the positive class is more poorly represented than the negative class. So we needed to use a weighted solution for linear ridge regression \cite{weighted_lrr}, in order to build a balanced classifier for our "part of the object vs others/context" classifiers.

We prove that for a specific form of weights, the weighting can be applied after computing the simple version (closed form linear regression, Eq. \ref{eq:simplified_lin_ridge_reg}). This is very important for our algorithm, because for the simple ridge regression we need to compute only one matrix inverse for all classifiers in one step, one matrix that all of them will share: ($\mathbf{D^\top D} +\lambda \mathbf{I}_k)^{-1}$ from Eq. \ref{eq:simplified_lin_ridge_reg}. For the weighted case, the closed form solution (as in  \cite{weighted_lrr}) would be different from classifier to classifier:

\begin{equation}
    \label{eq:weight_lin_ridge_reg}
    (\mathbf{D^\top W}_i \mathbf{D} +\lambda \mathbf{I}_k)\mathbf{\theta}_i= \mathbf{D^\top W}_i \mathbf{y}_i
\end{equation}

The weight matrix for a classifier $\mathbf{W_i}$ has the following form:
\begin{equation}
    \label{eq:sparse_w}
    \mathbf{W_i} = \mathbf{I}_n + 
    \begin{bmatrix}
            0 & 0 & 0 & ... & 0 & 0 & 0 \\
            0 & 0 & 0 & ... & 0 & 0 & 0 \\
            ...\\
            0  & 0 & ... & w_i &... & 0 & 0\\
            ...\\
            0 & 0 & 0 & ... & 0 & 0 & 0 \\
    \end{bmatrix}
    = \mathbf{I}_n + \mathbf{W}_{sparse_i}
\end{equation}
with $0$s and  $w_i$ only on $i^{th}$ position on the diagonal, $i$ being the index of the positive patch in data matrix, $\mathbf{D}$.

Replacing Eq.\ref{eq:sparse_w} in Eq.\ref{eq:weight_lin_ridge_reg}, and observing that $\mathbf{D^\top} \mathbf{W}_{sparse_i} = w_i [0|\mathbf{d}_i|0]$, the right hand side becomes: $\mathbf{D^\top W}_i \mathbf{y}_i = \mathbf{D^\top} (\mathbf{I}_n + \mathbf{W}_{sparse_i}) \mathbf{y}_i = \mathbf{D^\top} \mathbf{y}_i + w_i [0| \mathbf{d}_i | 0] \mathbf{y}_i = \mathbf{d}_i + w_i \mathbf{d}_i $. So, for the right term we get:

\begin{equation}
    \label{eq:simplified_y}
    \mathbf{D^\top W}_i \mathbf{y}_i  = (1 + w_i) \mathbf{d}_i 
\end{equation}

By doing the same operations on the left term:
$\mathbf{D^\top W}_i \mathbf{D} = \mathbf{D^\top} (\mathbf{I}_n + \mathbf{W}_{sparse_i}) \mathbf {D} = \mathbf{D^\top D} + w_i [0|\mathbf{d}_i|0] \mathbf{D} = \mathbf{D^\top D} + w_i \mathbf{d}_i  \mathbf{d}_i^\top$, the Eq. \ref{eq:weight_lin_ridge_reg} can be rewritten:

\begin{equation}
    \label{eq:simplified_weight_lin_ridge_reg}
      (\mathbf{D^\top D} + w_i \mathbf{d}_i  \mathbf{d}_i^\top +\lambda \mathbf{I}_k)\mathbf{\theta}_i= (1 + w_i) \mathbf{d}_i
\end{equation}

Let $\theta_i = q_i \mathbf{c}_i$, where $\mathbf{c}_i$ is the solution for linear ridge regression (Eq. \ref{eq:simplified_lin_ridge_reg}) and $q_i \in \mathbb{R}$. Then Eq. \ref{eq:simplified_weight_lin_ridge_reg} becomes:
$(\mathbf{D^\top D} + w_i \mathbf{d}_i \mathbf{d}_i^\top+\lambda \mathbf{I}_k) q_i \mathbf{c}_i= (1+w_i)\mathbf{d}_i$. From Eq. \ref{eq:simplified_lin_ridge_reg}, by simplifying terms we obtain
$q_i \mathbf{d}_i + q_i w_i \mathbf{d}_i \mathbf{d}_i ^\top \mathbf{c}_i = (1+w_i)\mathbf{d}_i$. Then, by multiplying at left with $\frac{ \mathbf{d}_i^\top}{||\mathbf{d}_i ||_2^2}$, we get:

\begin{equation}
    \label{eq:pre_q}
      q_i + q_i w_i  \mathbf{d}_i^\top \mathbf{c}_i = (1 + w_i)
\end{equation}

So, the solution for $q_i$ is ($w_i$ is $n-1$, because in "one vs all" classification, all elements in $\mathbf{D}$ are negative samples, except for one, the $i^{th}$):

\begin{equation}
    \label{eq:final_q}
      q_i  = \frac{(1 + w_i)} {1 + w_i  \mathbf{d}_i^\top \mathbf{c}_i} = \frac{n} {1 + (n-1)  \mathbf{d}_i^\top \mathbf{c}_i}
\end{equation}

Therefore, we proved that if $\mathbf{c}_i$ is the unique solution of linear ridge regression (since $\mathbf{D^\top D} + \lambda \mathbf{I}_k $ is always invertible, the solution in Eq. \ref{eq:simplified_y} is unique), then $q_i \mathbf{c}_i$ ($q_i$ from Eq. \ref{eq:final_q}) is the unique solution of Eq. \ref{eq:simplified_lin_ridge_reg} ($\mathbf{D^\top D} + w_i \mathbf{d}_i \mathbf{d}_i^\top+\lambda \mathbf{I}_k$ is always invertible, since it is also positive definite).

\section{Faster solution with efficient matrix inversion}
\label{subsec:matrix_inversion_lemma}
 Consider a general partitioned matrix $\mathbf{M}$
=
$\begin{matrix}
    \mathbf{E} & \mathbf{F} \\
	\mathbf{G} & \mathbf{H} \\
\end{matrix}$, with $\mathbf{E}$ and $\mathbf{H}$ invertible (Matrix Inversion Lemma~\cite{DBLP:books/lib/Murphy12}, Ch. 4.3.4.2). Then the following relation takes place:
 
 \begin{equation}
   (\mathbf{E} - \mathbf{F H}^{-1}\mathbf{G})^{-1} \mathbf{F} \mathbf{H}^{-1} = \mathbf{E}^{-1}\mathbf{F}(\mathbf{H} - \mathbf{G}\mathbf{E}^{-1}\mathbf{F})^{-1}
\end{equation}

By making the replacement: $\mathbf{E} = \lambda \mathbf{I}_k$, $\mathbf{H} =  \mathbf{I}_n$, $\mathbf{F} = \mathbf{D}^\top $, $\mathbf{G} = -\mathbf{D}$ ($\mathbf{E}$ and $\mathbf{H}$ are invertible) and rearranging the terms, we obtain ~\cite{DBLP:books/lib/Murphy12} (Ch. 14.4.3.2):
 
 \begin{equation}
    \label{eq:matrix_invers_lin_regres}
    (\mathbf{D^\top D} +\lambda \mathbf{I}_k)^{-1} \mathbf{D^\top} =  \mathbf{D^\top} (\mathbf{D D^\top} +\lambda \mathbf{I}_n)^{-1}
\end{equation}
We observe that the first term in Eq. \ref{eq:matrix_invers_lin_regres} is part of the closed form solution for the linear regression (without labels $y_i$). So, we can replace it with the one easier to compute. Since the bottleneck here is inverting the positive definite matrix $\mathbf{D^\top D} +\lambda \mathbf{I}_k$ or $\mathbf{D D^\top} +\lambda \mathbf{I}_n$, we will choose the easiest to invert. And this is the smaller one. In our case, $n$ is the number of patches, and $k$ is the number of features in each patch (equal to the number of pixels in the patch $\times$ number of pixel level channels, which is 7). A roughly approximation for $n$ is 500 and approximations for $k$ are $2000 \approx 17 \times 17 \times 7 $, $5000 \approx 27 \times 27 \times 7 $ and bigger for patches of bounding box size.

The second solution for computing the classifier is inverting a matrix two orders of magnitude smaller (as number of elements) then the first solution. So we choose the second part of Eq. \ref{eq:matrix_invers_lin_regres} for the closed form solution.

\end{appendices}

\end{document}